\documentclass[lettersize,journal]{IEEEtran}
\usepackage{amsmath,amsfonts}

\usepackage[linesnumbered, ruled]{algorithm2e}
\SetKwRepeat{Do}{do}{while}%
\usepackage{cite}
\usepackage{array}
\usepackage[caption=false,font=normalsize,labelfont=sf,textfont=sf]{subfig}
\usepackage{textcomp}
\usepackage{stfloats}
\usepackage{url}
\usepackage{verbatim}
\usepackage{graphicx}
\usepackage[T1]{fontenc} 
\usepackage{amssymb} 
\usepackage[backref=False]{hyperref}
\def\BibTeX{{\rm B\kern-.05em{\sc i\kern-.025em b}\kern-.08em
    T\kern-.1667em\lower.7ex\hbox{E}\kern-.125emX}}
\usepackage{balance}
\usepackage{soul}
\newcommand{\Rmnum}[1]{\uppercase\expandafter{\romannumeral #1}}  
\usepackage{tikz,xcolor,hyperref}
\usepackage{booktabs}
\definecolor{lime}{HTML}{A6CE39}
\usepackage{amsthm}
\newtheorem{remark}{Remark}

\begin{document}
\title{GAN-Based Semantic Communication for Image Transmission in IoV}
\author{Ruixing~Ren, Shan~Chen, Junhui~Zhao,~\IEEEmembership{Senior~Member,~IEEE,} Xiaoke~Sun

}
\maketitle

\begin{abstract}
For cooperative perception in the internet of vehicles, this paper proposes a generative adversarial network-based semantic communication framework to address the efficiency and fidelity bottlenecks of traditional communication systems in visual data transmission under limited bandwidth and dynamic channel conditions. At the transmitter, the framework adopts a pyramid attention network to extract semantic label maps and introduces a semantic priority preservation mechanism. It assigns differentiated weights to distinct semantic categories based on driving safety, guiding bit allocation and loss function design. At the receiver, an image reconstruction module integrating a coarse-to-fine multi-resolution generator and multi-scale discriminator is designed. Combined with the temporal consistency branch, spatial pyramid pooling and class-aware convolutional layers, it achieves high-fidelity reconstruction of high-quality images from corrupted semantic labels. The model is trained with combined adversarial, feature matching and perceptual losses, effectively improving semantic consistency and visual realism of generated images. Experimental results on the Cityscapes dataset show that the proposed method outperforms existing counterparts in both semantic segmentation accuracy and reconstructed image quality, and maintains stable reconstruction performance under AWGN and Rayleigh channels.
\end{abstract}


\section{Introduction}
With the rapid development of intelligent transportation systems, the internet of vehicles (IoV) faces increasingly severe data communication challenges \cite{RenUAV, RenDCAN}. Vehicles need to share massive visual information in real time to support critical tasks such as cooperative perception and safe driving \cite{RenITS}. However, spectrum scarcity and network congestion become increasingly prominent. Achieving efficient and reliable visual data transmission under limited bandwidth and highly dynamic channels has become a critical issue in IoV research.

Semantic communication (SC) has attracted widespread attention as an emerging paradigm beyond conventional bit-level transmission. Unlike syntax-level communication, it aims to extract and transmit inherent semantic meaning rather than complete data waveforms. This drastically reduces data transmission volume, improves communication efficiency, and maintains favorable semantic fidelity under harsh channels, showing great potential in resource-limited and task-driven scenarios such as IoV.

In recent years, remarkable progress has been made in SC. Early works by Bao et al. \cite{Bao} explored the theoretical relationship between SC and Shannon information theory. Guler \cite{Guler} investigated semantic information transmission under agent assistance. Xie et al. \cite{Xie} developed the deep learning-based SC system DeepSC for text transmission. Weng \cite{Weng} attempted to improve the recovery quality of speech signals in SC systems by reducing semantic-level errors rather than bit- or symbol-level errors. These studies reveal that the core of SC lies in efficiently extracting semantic information from raw data, and recovering task-compliant images with high quality from corrupted semantic features at the receiver. Especially, maintaining image semantic fidelity under low bit rate and high noise conditions becomes a critical bottleneck restricting the application of SC in visual tasks. One feasible solution to address this challenge is to develop powerful generative models.

Among various generative models, generative adversarial networks (GANs) exhibit remarkable advantages in enhancing semantic feature extraction and reconstruction quality within SC systems. Through adversarial training between generators and discriminators, GANs can produce high-fidelity samples, which is critical for the efficient compression and recovery of semantic information. The original GAN proposed by Goodfellow et al. \cite{Goodfellow} achieved breakthrough progress in image generation. Subsequently, conditional GAN \cite{cGAN} and its variants have performed excellently in image-to-image translation tasks. In SC, researchers have applied GANs to optimize the design of semantic encoders and decoders \cite{Zhu,Han}. However, efficiently integrating GANs with semantic feature extraction and reconstruction in IoV remains an open issue.

Unlike general image communication, visual data transmission in IoV faces its own unique challenges, which directly motivate the design of this paper. First, high vehicle mobility and rapidly changing scenes make the visual content vary drastically within a short time, so the communication system must preserve task-relevant semantics rather than pixel-level details under limited bandwidth \cite{RenRIS}. Second, cooperative perception is subject to strict end-to-end latency constraints, leaving no room for retransmitting massive visual data; the transmitted representation must therefore be extremely compact. Third, driving scenes contain safety-critical objects (e.g., pedestrians, vehicles, and traffic signs) whose misinterpretation may lead to severe accidents, so unequal protection according to semantic importance is indispensable. Fourth, consecutive frames captured by a moving vehicle exhibit strong temporal correlation, and ignoring it leads to flickering and temporal inconsistency in the reconstructed sequence. Fifth, vehicular wireless channels are highly dynamic due to fading, shadowing, and interference, demanding a semantic representation that is robust against channel impairments. The proposed framework responds to these challenges respectively: the semantic label map provides a compact and robust representation; the semantic priority preservation mechanism safeguards safety-critical categories; the temporal consistency branch exploits inter-frame correlation; and the GAN-based decoder recovers high-fidelity images under low bit rates and harsh channels.

Based on existing GAN technologies, this paper further optimizes the generation and discrimination mechanisms in SC systems to achieve more effective semantic information transmission and reconstruction in IoV. The main contributions of this paper are as follows:
\begin{itemize}
	\item A semantic priority preservation mechanism is designed at the encoder, assigning smaller quantization step sizes and higher channel coding redundancy to high-priority regions. Different penalty weights are imposed on various categories in the segmentation loss function, forcing the network to prioritize the classification accuracy of critical regions.
	\item A two-stage cascaded generator is designed to restore global layout and texture details step by step. A multi-scale discriminator is adopted; meanwhile, temporal consistency branch, spatial pyramid pooling and class-aware convolutional layers are introduced for IoV, enabling high-fidelity reconstruction from sparse semantic labels.
	\item Experiments are conducted on the Cityscapes dataset to evaluate semantic segmentation accuracy, reconstructed image quality, and robustness under AWGN and Rayleigh channels, which verifies the effectiveness of the proposed method.
\end{itemize}

Section \ref{2} introduces the system model.
Section \ref{3} elaborates on the proposed scheme in detail.
Section \ref{4} presents the simulation experiments and result analysis.
Section \ref{5} concludes the whole paper.

\section{System Model}\label{2}
Consider the cooperative perception scenario in IoV. Vehicles capture road environment images via onboard cameras and transmit them to other vehicles through wireless channels \cite{RenIoV}. Let the raw image be $\mathbf{I}\in\mathbb{R}^{C\times H\times W}$, where $C$, $H$, and $W$ denote the channel number, height, and width of the image, respectively. The GAN-based SC system can be decomposed into two stages: the encoder extracts the semantic label map and transmits it after channel coding; the decoder recovers the semantic label map and reconstructs the image.

At the transmitter, the semantic encoder is denoted as $\mathcal{P}$, which maps the raw image into a semantic label map $\mathbf{S}=\mathcal{P}(\mathbf{I}) \in\mathcal{C}^{H\times W}$. Here, $\mathcal{C}=\{1,2,\dots,M\}$ represents $M$ predefined semantic categories, such as road, vehicle, and pedestrian. The channel encoder is denoted by $\mathcal{W}_{\text{enc}}$, which encodes the semantic label map into a latent representation with length $n$, i.e., $\mathbf{z}=\mathcal{W}_{\text{enc}}(\mathbf{S}) \in\mathbb{R}^n.$

The wireless channel is modeled as $\mathcal{H}(\cdot)$, and the received signal at the receiver is expressed as:
$\tilde{\mathbf{z}} = \mathcal{H}(\mathbf{z})$. For the AWGN channel, it can be further written as $\tilde{\mathbf{z}}=\mathbf{z}+\mathbf{n}$, where $\mathbf{n}\sim\mathcal{N}(\mathbf{0},\sigma^2\mathbf{I})$. For the Rayleigh fading channel, the received signal is expressed as $\tilde{\mathbf{z}} = \mathbf{h} \odot \mathbf{z} + \mathbf{n}$, where $\mathbf{h}$ denotes the fading coefficient vector.  Define the channel decoder as $\mathcal{W}_\text{dec}$, which recovers the semantic label map as
$\hat{\mathbf{S}}=\mathcal{W}_{\text{dec}}(\tilde{\mathbf{z}}) \in \mathcal{C}^{H\times W}.$ Generally, $\mathcal{W}_{\text{dec}}$ acts as the inverse process of $\mathcal{W}_{\text{enc}}$. Let the image generator be denoted as $G$; then the GAN-based reconstructed image is given by $\hat{\mathbf{I}}=G(\hat{\mathbf{S}}) \in\mathbb{R}^{C\times H\times W}$. In summary, the overall system procedure can be formulated as
\begin{equation}
	\hat{\mathbf{I}}=G(\mathcal{W}_{\text{dec}}(\mathcal{H}(\mathcal{W}_{\text{enc}}(\mathcal{P}(\mathbf{I}))))).
\end{equation}

\begin{remark}
	The real-valued model above is an equivalent representation of a complex baseband system: separating the in-phase and quadrature components maps $n/2$ complex dimensions to $n$ real ones, the per-dimension real AWGN corresponds to circularly symmetric complex AWGN, and Rayleigh fading scales both quadratures, preserving the element-wise form $\tilde{\mathbf{z}}=\mathbf{h}\odot\mathbf{z}+\mathbf{n}$. In the simulated chain, the label map ($\lceil\log_2 M\rceil$ bits/pixel before protection) passes through priority-aware bit allocation, channel coding, and modulation into $\mathbf{z}$, and the reported bpp counts all transmitted bits, including coding redundancy, per pixel. The fading coefficients are i.i.d.\ Rayleigh per symbol, and the receiver assumes perfect CSI and equalizes before decoding.
\end{remark}

The objective of this paper is to minimize the semantic distortion between the original image $\mathbf{I}$ and the reconstructed image $\hat{\mathbf{I}}$, rather than the conventional pixel-level distortion.

\section{Proposed Scheme}\label{3}
The proposed scheme consists of two major modules: a semantic segmentation encoder at the transmitter and a semantics-aware image reconstruction decoder at the receiver.

\subsection{Semantic Segmentation Encoder}

\textbf{Semantic label map as a semantic representation.} For readers from the communication community, we briefly clarify the semantic label map. Semantic segmentation assigns each pixel a discrete category label (e.g., road, vehicle, pedestrian), yielding a pixel-wise class map $\mathbf{S}$ that abstracts the scene. Unlike raw images (pixel intensities, textures, colors) or learned feature maps (continuous, high-dimensional, entangled), the label map is discrete, low-dimensional, and interpretable, retaining only task-relevant structure. Its low entropy allows transmission with very few bits, and its discrete nature permits unequal protection by classical channel codes; hence it is a suitable semantic representation for task-oriented communication, with missing visual details regenerated by the generative decoder.

\textbf{Comparison with intermediate-feature transmission.} DJSCC-style methods \cite{DJSCC, DJSCCf} instead transmit learned continuous intermediate features. Such features achieve good pixel-level rate--distortion performance but are entangled and uninterpretable, making unequal protection of safety-critical categories difficult, and they remain sensitive to channel noise since they bypass discrete channel coding. In contrast, the discrete label map sacrifices texture/color details but gains interpretability, compactness, and compatibility with priority-aware quantization and coding; the discarded details are unnecessary for the perception task and are regenerated by the GAN decoder.

\textbf{Basic segmentation network.} To extract high-dimensional semantic information from the raw image $\mathbf{I}$, this paper adopts the pyramid attention network (PAN) \cite{PAN} as $\mathcal{P}$. Taking ResNet-50 as the backbone, PAN first extracts feature maps, and then outputs the final semantic label map $\mathbf{S}$ via the feature pyramid attention (FPA) module and the global attention upsampling (GAU) module.

\textbf{Semantic priority preservation mechanism.} Different semantic objects differ greatly in importance to driving safety. Misdiagnosis or missed detection of pedestrians and vehicles may trigger severe traffic accidents, while degraded reconstruction quality of sky and buildings barely affects decision-making. Therefore, a semantic priority preservation mechanism is introduced to allocate more bits and enhanced protection to critical categories during encoding and transmission.

The $M$ semantic categories are divided into $K$ priority levels according to their importance to driving safety, where $\alpha^{(k)}$ denotes the weight coefficient of the $k$-th level. For image position $(i,j)$, $P(c|I_{i,j})$ denotes the probability of belonging to category $c$ output by the segmentation network. The importance weight of position $(i,j)$ is defined as:
\begin{equation}
	\omega_{i,j}=\sum_{c\in \mathcal{C}}\alpha^{(r(c))}\cdot P(c|I_{i,j}),
\end{equation}
where $r(c)$ represents the priority level of category $c$. This weight directly guides bit allocation in the channel encoder: regions with higher weights adopt smaller quantization steps and higher channel coding redundancy. In our implementation, we set $K=3$ priority levels: the highest level ($\alpha^{(1)}=1.0$) comprises safety-critical categories (person, rider, vehicle, traffic light, and traffic sign); the second level ($\alpha^{(2)}=0.8$) comprises navigation-relevant categories (road and sidewalk); and the third level ($\alpha^{(3)}=0.5$) comprises background categories (building, wall, fence, pole, vegetation, terrain, and sky). The level index determines the source and channel protection: higher-priority regions are quantized with smaller steps and encoded with lower-rate channel codes (i.e., more redundancy), whereas lower-priority regions use coarser steps and higher-rate codes; equivalently, the quantization step and the code rate increase monotonically as $\alpha^{(k)}$ decreases.

We further impose priority weighting on the semantic segmentation loss in the training loss function. Let $D(S_{i,j},\hat{S}_{i,j})$ denote the difference between the original semantic label map $\mathbf{S}$ and the reconstructed label map $\hat{\mathbf{S}}$ at position $(i,j)$. Specifically, $D(\cdot,\cdot)$ is the cross-entropy between the one-hot ground-truth label at $(i,j)$ and the class-probability vector produced by the decoder, i.e., $D(S_{i,j},\hat{S}_{i,j})=-\log \hat{S}_{i,j}(S_{i,j})$, making (3) a priority-weighted cross-entropy. The weighted semantic loss is formulated as:
\begin{equation}
	\mathcal{L}_{\text{sem}}=\sum_{i,j}\omega_{i,j} \cdot D(S_{i,j},\hat{S}_{i,j}).
\end{equation}
This loss drives the network to prioritize classification accuracy in high-priority regions. With the coefficient values specified above, empirical results verify that the priority weighting effectively improves the reconstruction fidelity of safety-critical regions.

\textbf{Segmentation performance evaluation metrics.} Intersection over Union (IoU) and mean IoU (mIoU) are adopted to evaluate segmentation quality. For category $m$, let $\mathbf{G}_m$ denote the ground-truth pixel set and $\mathbf{P}_m$ the prediction set, where
\begin{equation}
	\text{IoU}_m = \frac{\left| \mathbf{G}_m\cap \mathbf{P}_m \right|}{\left| \mathbf{G}_m\cup \mathbf{P}_m \right|},
\end{equation}
\begin{equation}
	\text{mIoU}=\frac{1}{M}\sum_{m=1}^{M}\text{IoU}_m.
\end{equation}
The optimization objective of semantic segmentation is to maximize mIoU, which is equivalent to minimizing the loss $\mathcal{L}_{\text{seg}}=1-\text{mIoU}$.

\begin{figure}[t]
	\centerline{\includegraphics[width=3.5in,keepaspectratio]{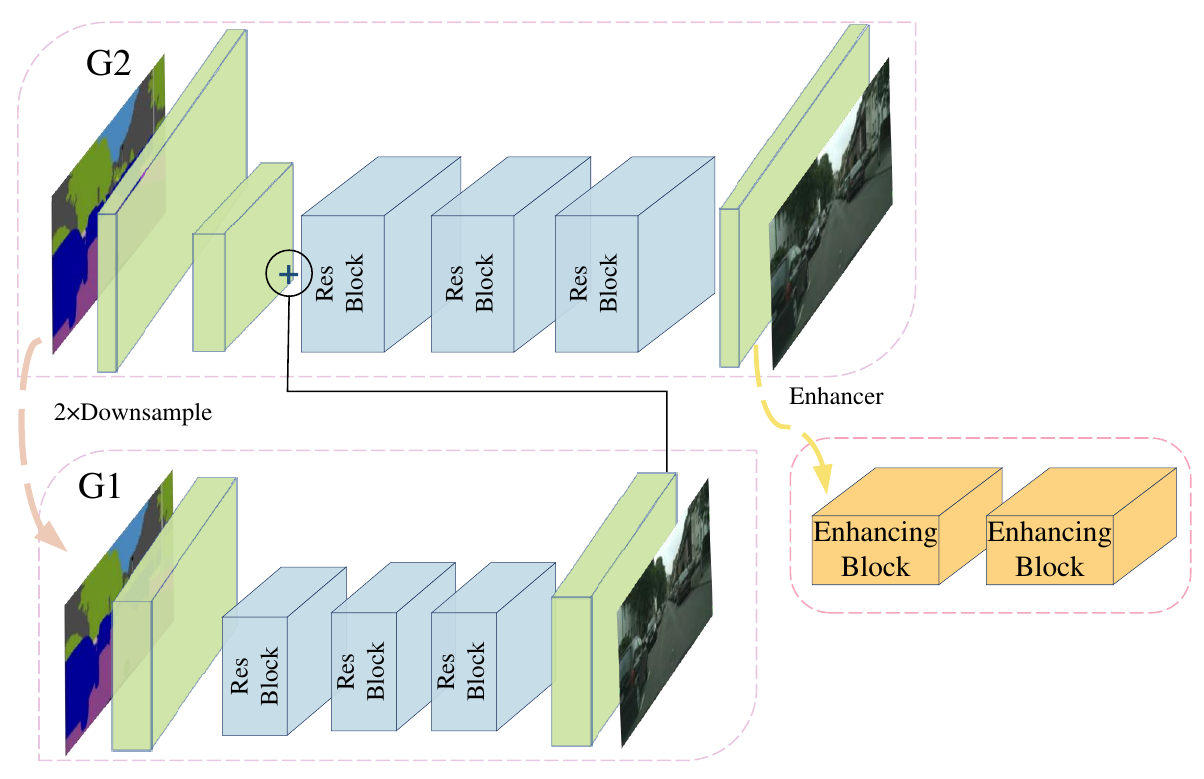}}
	\caption{Generator Network Architecture.}
	\label{Fig3}
\end{figure}

\subsection{Image Reconstruction Decoder}
The receiver reconstructs the photorealistic image $\hat{\mathbf{I}}$ from the recovered semantic label map $\hat{\mathbf{S}}$.

\textbf{Coarse-to-fine multi-resolution generator.} A single generator hardly captures global structures (e.g., road layout and sky contour) and local details (e.g., text on traffic signs) simultaneously. High-resolution input greatly increases computational overhead, while low-resolution input leads to detail loss. The design logic is therefore as follows: a single-stage generator must trade off receptive field against resolution and cannot recover the global layout and the local details simultaneously; accordingly, the first stage restores the global scene layout from a low-resolution input with a large receptive field, and the second stage refines local texture and edge details at full resolution, which also keeps the computational cost compatible with efficient IoV deployment. Therefore, as illustrated in Figure \ref{Fig3}, we decompose the generator into two cascaded subnetworks $G=\{G_1,G_2\}$ with a coarse-to-fine strategy:
\begin{itemize}
	\item Global generator $G_1$: It takes the $2\times$ downsampled semantic map as input and outputs a low-resolution image ($1024\times512$) to reconstruct the overall scene layout. Equipped with larger $5\times5$ convolutional kernels, it achieves a larger receptive field.
	\item Local enhancer $G_2$: It concatenates the output of $G_1$ with its front-end features to refine details at the original resolution, finally producing a $2048\times1024$ output that can be further extended to $4096\times2048$ after enhancement. With smaller $3\times3$ kernels, it focuses on texture and edge reconstruction.
\end{itemize}
The two generators share an identical structure, consisting of a front convolution block $G^{(F)}$, three residual blocks $G^{(R)}$, and a rear deconvolution block $G^{(B)}$. $G_2$ additionally incorporates two $3\times3$ convolutional layers at the rear to enhance high-frequency details. This design achieves a favorable balance between computational cost and reconstruction quality. Furthermore, a two-stage training strategy is adopted: we first train $G_1$, then fine-tune $G_2$ jointly, and finally train the generator and discriminator in an end-to-end manner. For reproducibility, we summarize the complete training procedure. The segmentation encoder and the reconstruction network are trained separately: PAN is first pre-trained with the priority-weighted semantic loss (3) and then fixed, and its predicted label maps are used to train the reconstruction module. Within the reconstruction module, $G_1$ is trained first, $G_2$ is then fine-tuned jointly, and finally the generator and the discriminators are optimized end-to-end with the adversarial, feature-matching, and perceptual losses, where the generator's parameters are held fixed when updating each discriminator. The semantic priority loss (3) is thus applied at the segmentation pre-training stage, while the priority weights additionally guide bit allocation and channel protection at the transmitter.

\textbf{Computational complexity.} On an NVIDIA RTX 4060 Ti with batch size 1 and $512\times1024$ input, the generator has 182.6 M parameters (about 696 MB in float32) and 605.6 G FLOPs, with an average inference latency of 94.5 ms (about 10.6 FPS). We report these figures as an explicit computational-complexity analysis of the reconstruction module, rather than as a real-time claim.

\textbf{Multi-scale discriminator.} A single discriminator cannot effectively evaluate global structure and local texture simultaneously. Therefore, two discriminators $D_1$ and $D_2$ are employed to process the original-resolution image and the $2\times$ downsampled image, respectively. $D_1$ focuses on fine-grained details (e.g., wheel spokes and traffic signal text), while $D_2$ evaluates global consistency (e.g., road continuity and vehicle proportion). The two discriminators jointly provide gradient feedback to the generator, substantially improving reconstruction realism.

To adapt to the dynamic characteristics of IoV scenarios, we further improve the discriminator:
\begin{itemize}
	\item Temporal consistency branch (embedded in \(D_1\)): Considering the strong temporal correlation between adjacent frames in IoV, single-frame reconstruction is prone to flickering artifacts. We compute feature correlation between the current and historical frames to alleviate jitter and blurring under high-speed movement. Formally, let $F_t$ and $F_{t-1}$ be the intermediate feature maps extracted by $D_1$ from the current and historical reconstructed frames; the per-location temporal correlation map is $c_{u,v}=\langle F_t^{u,v},F_{t-1}^{u,v}\rangle/(\|F_t^{u,v}\|\,\|F_{t-1}^{u,v}\|)$, which modulates the adversarial gradient of $D_1$ and enters the generator objective as the temporal regularizer $\mathcal{L}_{\text{tc}}=\mathbb{E}\,\|c\odot(F_t-F_{t-1})\|_1$. High-correlation regions are thus enforced to remain temporally consistent, suppressing flicker, while low-correlation regions (occlusions, fast motion) rely on the current frame, avoiding motion blur from misalignment.
	\item Scale-adaptive spatial pyramid pooling: Objects captured by onboard cameras exhibit large distance variations, which cannot be well covered by a fixed receptive field. We extract features using pooling windows of \(1\times1\), \(2\times2\), and \(4\times4\) to represent distant, medium, and near objects.
	\item Class-aware convolutional layers: Safety-critical regions require higher reconstruction priority, and the discriminator is designed to guide the generator to optimize these regions preferentially. Higher discrimination weights are assigned to key categories such as pedestrians and vehicles.
\end{itemize}
The conditional adversarial loss of a single discriminator is defined as:
\begin{equation}
	\begin{aligned}
		\mathcal{L}_{\text{adv}}(D_k, G) &= \mathbb{E}_{(\hat{\mathbf{S}}, \mathbf{I})}\left[\log D_k(\hat{\mathbf{S}}, \mathbf{I})\right] \\
		&+ \mathbb{E}_{\hat{\mathbf{S}}}\left[\log\left(1 - D_k\left(\hat{\mathbf{S}}, G(\hat{\mathbf{S}})\right)\right)\right].
	\end{aligned}
\end{equation}
For the dual discriminator, the total adversarial loss is formulated as:
\begin{equation}
	\min_G\max_{D_1,D_2}\sum_{k=1,2}\mathcal{L}_{\text{adv}}(D_k,G).
\end{equation}

\textbf{Loss function design.} Relying solely on adversarial loss tends to cause mode collapse and artifacts. Therefore, we introduce two auxiliary loss functions. First, we enforce the generated images to be close to the real images at the intermediate feature layers of the discriminator, which is defined as the feature matching loss \(\mathcal{L}_{\text{fm}}\). Let \(D_k^{(i)}\) denote the \(i\)-th feature layer of discriminator \(D_k\), and \(T_i\) be the number of elements in the feature map. Then:
\begin{equation}
	\mathcal{L}_{\text{fm}}(G, D_k)
	= \mathbb{E}_{(\hat{\mathbf{S}}, \mathbf{I})}
	\sum_{i=1}^{N} \frac{1}{T_i}
	\left\|
	D_k^{(i)}(\hat{\mathbf{S}}, \mathbf{I}) - D_k^{(i)}(\hat{\mathbf{S}}, G(\hat{\mathbf{S}}))
	\right\|_1 .
\end{equation}
where \(N\) denotes the total number of layers. A pre-trained VGG network is further adopted to extract high-level features of real and generated images, and the L1 distance is calculated to define the perceptual loss \(\mathcal{L}_{\text{perc}}\). This loss focuses more on semantic content rather than pixel-level exact matching:
\begin{equation}
	\mathcal{L}_{\text{perc}} = \mathbb{E}_{(\mathbf{I},\hat{\mathbf{S}})} \sum_{i=1}^{N} \frac{1}{T_i} \left\| F^{(i)}(\mathbf{I}) - F^{(i)}(G(\hat{\mathbf{S}})) \right\|_1,
\end{equation}
where \(F^{(i)}\) denotes the \(i\)-th layer feature of the pre-trained VGG network. The overall loss can be formulated as
\begin{equation}
	\begin{aligned}
		\min_{G} \Bigg(
		&\max_{D_1,D_2} \sum_{k=1}^{2} \mathcal{L}_{\text{GAN}}(D_k, G) \\
		&+ \lambda_{\text{FM}} \sum_{k=1}^{2} \mathcal{L}_{\text{FM}}(G, D_k) + \lambda_{\text{perc}} \mathcal{L}_{\text{perc}}
		\Bigg).
	\end{aligned}
\end{equation}
where \(\lambda_{\text{FM}}\) and \(\lambda_{\text{perc}}\) are hyperparameters used to balance the contribution of each loss term.

\begin{figure}[t]
	\centerline{\includegraphics[width=3.2in,keepaspectratio]{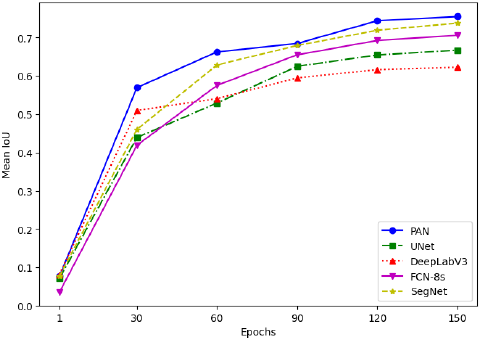}}
	\caption{mIoU for different SC models.}
	\label{Fig4}
\end{figure}

\section{Simulation Experiments and Analysis}\label{4}
The proposed system is trained and evaluated on the Cityscapes dataset. The Cityscapes dataset \cite{Cityscapes} contains 2975 training images and 1525 test images. In addition to mIoU and the accuracy of key categories as the main evaluation metrics, traditional image quality indices including PSNR and SSIM are also adopted for reference.

\textbf{Experimental setup.} The segmentation encoder is compared with UNet \cite{UNet}, FCN-8s \cite{FCN}, DeepLabV3 \cite{DeepLabV3}, and SegNet \cite{SegNet}, all trained on the same training split with their default settings (Figure~\ref{Fig4}). For reconstruction, the proposed method (labeled DL-ISC in the figure legends) is compared with JPEG \cite{JPEG}, JPEG2000 (J2K) \cite{J2K}, BPG \cite{BPG}, DSSLIC \cite{DSSLIC}, HiFiC \cite{HiFiC}, and CRN \cite{CRN} under their default codec configurations, at the same operating bit rates (0.1-0.75 bpp, where 0.1-0.3 bpp highlights extreme compression). All methods are evaluated on the same 1525 test images and the same channel realizations, with SNR from 0 to 5 dB in the robustness test, and the bit rate is computed consistently as defined in Remark 1, ensuring a fair bandwidth and channel comparison.

The pre-trained PAN is adopted as the semantic segmentation model to generate semantic labels for reconstructed images. Figure \ref{Fig4} presents the mIoU comparison of different models. As the number of iterations increases, PAN achieves a higher mIoU than UNet, FCN and other baseline models, which verifies the effectiveness of the proposed system in semantic segmentation and task-driven encoding.

\textbf{Ablation of the priority mechanism.} Table \ref{tab:priority_ablation} compares the priority-weighted variant with uniform protection ($\alpha^{(k)}\equiv 1$) under identical training settings. Because the weighting reallocates capacity from background categories toward safety-critical ones, the overall mIoU decreases slightly ($0.266\rightarrow0.257$) and the safety-class mIoU remains comparable ($0.259\rightarrow0.255$), while the most safety-critical vulnerable classes (person, rider, motorcycle) clearly improve from $0.181$ to $0.208$ ($+2.7$ percentage points; e.g., motorcycle $+5.2$, person $+1.6$, rider $+1.2$). This confirms that the mechanism shifts capacity toward the categories that matter most for driving safety as designed; its protection effect under bandwidth and channel constraints is further reflected in the robustness results of Figure~\ref{Fig9}.

\begin{table*}[t]
	\centering
	\caption{Ablation of the semantic priority preservation mechanism on the Cityscapes validation set (mIoU).}
	\label{tab:priority_ablation}
	\begin{tabular}{lccc}
		\toprule
		Variant & Overall mIoU & Safety-class mIoU & Vulnerable-class mIoU \\
		\midrule
		Uniform protection ($\alpha^{(k)}\equiv 1$) & 0.266 & 0.259 & 0.181 \\
		Priority weighting (proposed) & 0.257 & 0.255 & 0.208 \\
		\bottomrule
	\end{tabular}
\end{table*}

\begin{figure}[t]
	\centerline{\includegraphics[width=3.5in,keepaspectratio]{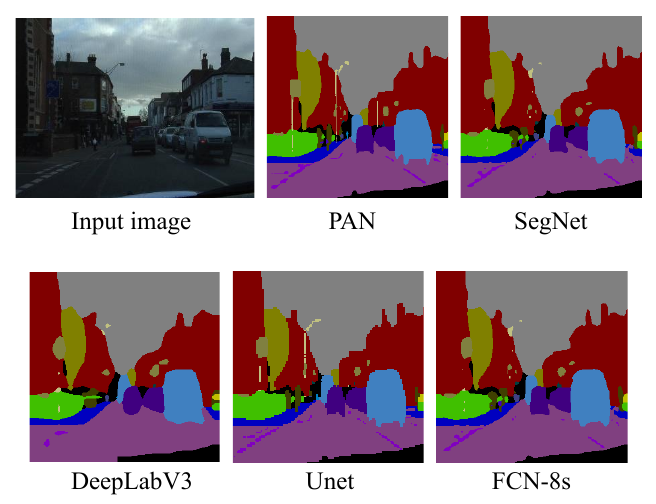}}
	\caption{Semantic Segmentation Encoding Example.}
	\label{Fig5}
\end{figure}

\begin{figure*}[t]
	\centerline{\includegraphics[width=6.2in,keepaspectratio]{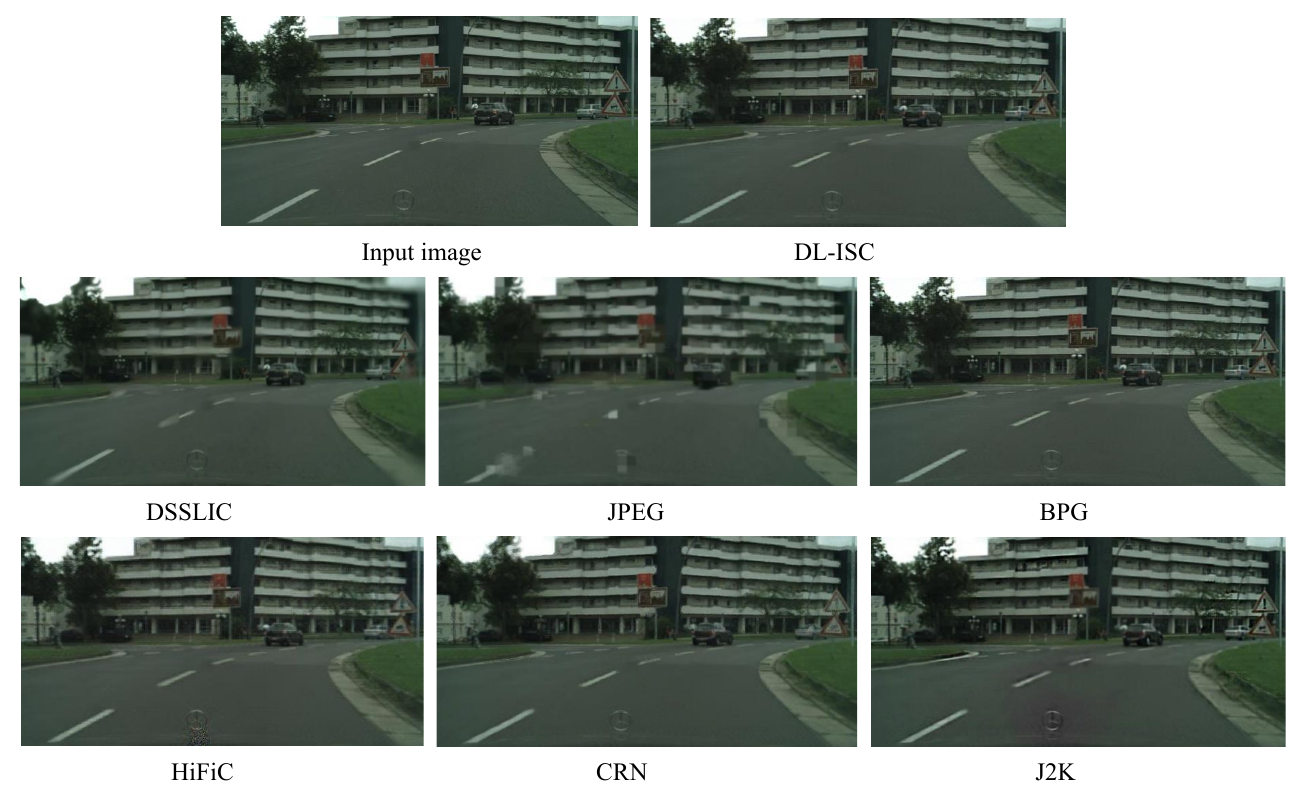}}
	\caption{Comparison of Reconstruction Performance Among Different Image Encoders.}
	\label{Fig6}
\end{figure*}

\begin{table*}[!]
	\centering
	\caption{GAN Network Architecture (Multi-Scale Discriminator)}
	\label{tab:gan_architecture}
	\begin{tabular}{lll}
		\toprule
		Discriminator & Layer & Operation \\
		\midrule
		$D_1$  & layer 0 & Conv2d(39$\rightarrow$64, 4$\times$4, stride=2), LeakyReLU \\
		& layer 1 & Conv2d(64$\rightarrow$128, stride=2), LeakyReLU \\
		& layer 2 & Conv2d(128$\rightarrow$256, stride=2), LeakyReLU \\
		& layer 3 & Conv2d(256$\rightarrow$512, stride=1), LeakyReLU \\
		& layer 4	 & Conv2d(512$\rightarrow$1, stride=1) \\
		\addlinespace
		$D_2$ & layer 0 & Conv2d(39$\rightarrow$64, 4$\times$4, stride=2), LeakyReLU \\
		& layer 1 & Conv2d(64$\rightarrow$128, stride=2), LeakyReLU \\
		& layer 2 & Conv2d(128$\rightarrow$256, stride=2), LeakyReLU \\
		& layer 3 & Conv2d(256$\rightarrow$512, stride=1), LeakyReLU \\
		& layer 4 & Conv2d(512$\rightarrow$1, stride=1) \\
		\bottomrule
	\end{tabular}
\end{table*}

Figure \ref{Fig5} shows the semantic label maps generated by different models. It can be observed that the segmentation results of PAN are more consistent with the real scene and possess stronger capability of object recognition and localization, providing a reliable foundation for subsequent image analysis. By contrast, other models are prone to block artifacts, blurring and ringing artifacts under low bit rates.

Figure \ref{Fig6} illustrates the reconstruction performance of different image encoders. The images reconstructed by the proposed method are highly similar to the inputs with favorable quality consistency, while other models are susceptible to blurring and block artifacts at low bit rates. GAN gradually improves the realism of generated images through adversarial training while preserving semantic information, in which the discriminator network plays a critical role. The detailed structure of the discriminator network is presented in Table \ref{tab:gan_architecture}. In Table \ref{tab:gan_architecture}, the 39 input channels comprise the 3 RGB channels of the image, the 35 one-hot channels of the semantic label map, and 1 instance-map channel, i.e., $3+35+1=39$, which are concatenated as the conditional input of the discriminators.

\begin{figure*}[t]
	\centerline{\includegraphics[width=7.2in,keepaspectratio]{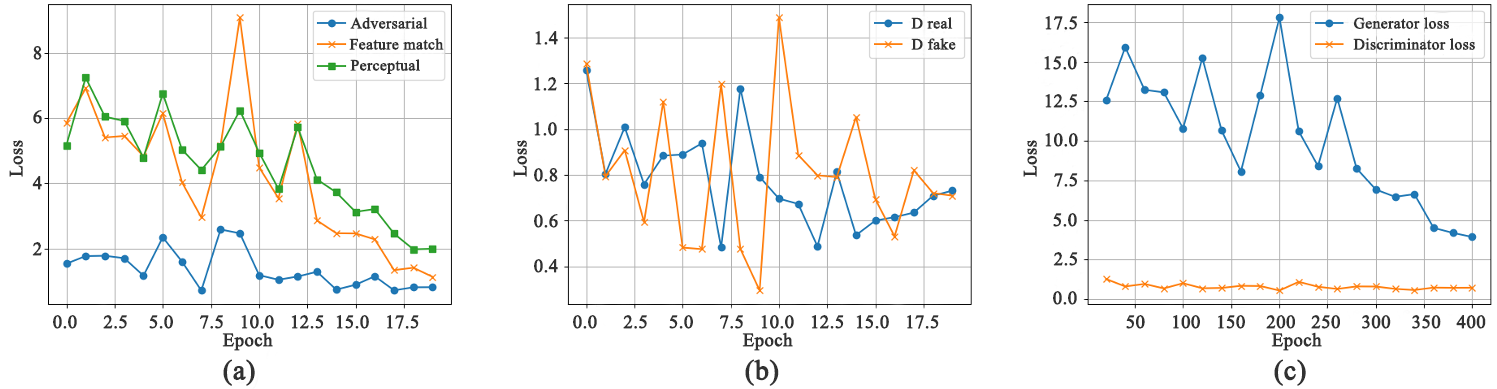}}
	\caption{Variation of Loss Function During Image Reconstruction.}
	\label{Fig7}
\end{figure*}
\begin{figure}[!]
	\centerline{\includegraphics[width=3.2in,keepaspectratio]{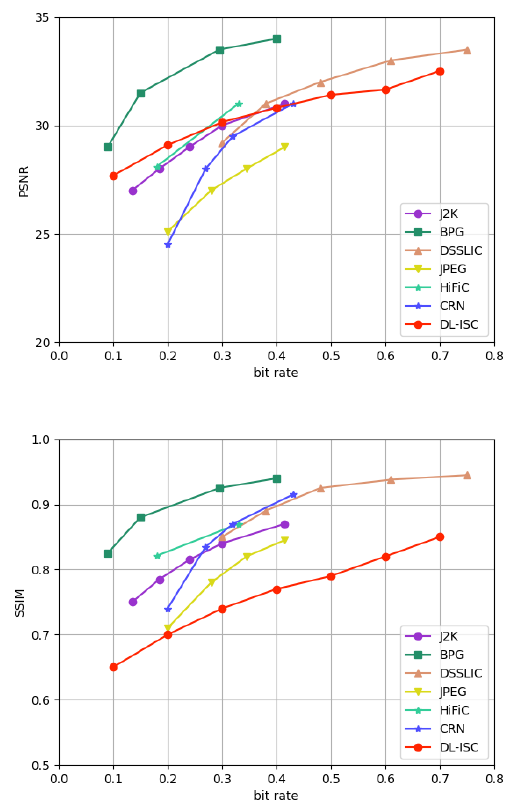}}
	\caption{Rate-Distortion Performance Comparison of Different Image Encoders}
	\label{Fig8}
\end{figure}

Figure \ref{Fig7}(a) depicts the iterative trends of the three generator losses. The adversarial loss declines slowly, reflecting the ongoing optimization of the generator-discriminator game. Both the feature matching loss and perceptual loss converge rapidly. The above results demonstrate that the proposed method can effectively guide the generator to approximate the distribution of real images simultaneously at the feature and semantic levels. Figure \ref{Fig7}(b) shows the discriminator loss, which comprises losses from both real and generated images. Balancing the weights of these two terms effectively enhances the  discriminator's capability to differentiate between real and fake samples. Figure \ref{Fig7}(c) shows the loss curves of the generator and discriminator. The generator loss exhibits a fluctuating downward trend, reflecting the continuous improvement of its generative performance. The discriminator loss, however, remains stable, indicating its sustained capability to distinguish between real and generated samples.
\begin{figure}[!]
	\centerline{\includegraphics[width=3.2in,keepaspectratio]{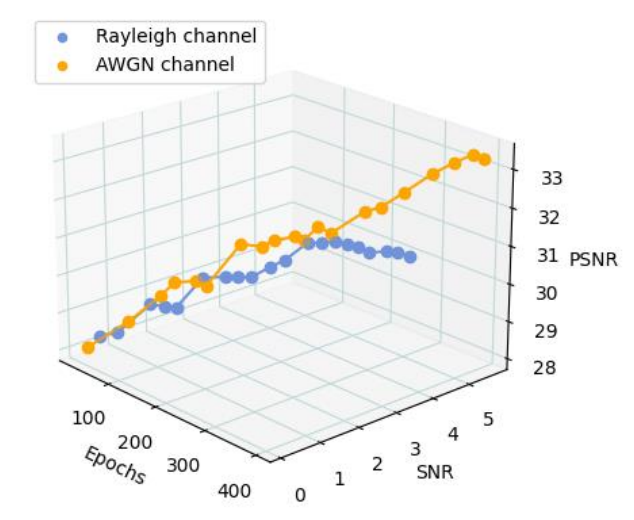}}
	\caption{PSNR-SNR Performance Under AWGN and Rayleigh Channels.}
	\label{Fig9}
\end{figure}

Figure \ref{Fig8} depicts the rate-distortion performance. In the low bit rate range of 0.1--0.3 bpp, the proposed method achieves a higher PSNR than several classical and learned schemes (e.g., JPEG, J2K, and CRN). This indicates that under extreme compression conditions, image reconstruction leveraging semantic features can better preserve visually perceivable information and further improve the robustness of the system. Across the evaluated bit-rate range, pixel-optimized codecs such as BPG achieve a higher PSNR than the proposed method, since PSNR rewards pixel-level fidelity that these codecs directly optimize; the benefit of the proposed method lies instead in semantic fidelity and robustness under low bit rates and channel impairments, which illustrates the tradeoff between pixel-level PSNR and semantic/perceptual quality. At extremely high bit rates, the fine details of GAN-reconstructed images are not yet comparable to those of traditional lossless compression, leading to a slight disadvantage in terms of the PSNR metric. Overall, the proposed method has a certain tolerance to pixel errors, and can maintain satisfactory visual quality even under low bit rates or channel interference.

Figure \ref{Fig9} evaluates the robustness under AWGN and Rayleigh fading channels. In the AWGN channel, the PSNR of reconstructed images gradually improves as the SNR increases. Even under low SNR conditions, the proposed scheme can reconstruct visually acceptable images by compensating for missing information. Compared with the AWGN channel, the Rayleigh channel introduces inherent fading and distortion, resulting in an overall lower PSNR curve. Even so, the proposed scheme can partially counteract the impact of channel fading and achieve more stable reconstruction quality.

\section{Conclusion}\label{5}
This paper proposed a GAN-based SC framework for IoV cooperative perception. At the transmitter, a pyramid attention network was adopted to extract semantic label maps, and a semantic priority preservation mechanism was designed to prioritize the integrity of key regions under bandwidth constraints. At the receiver, a coarse-to-fine image reconstruction module is constructed by integrating a multi-resolution generator and a multi-scale discriminator. The model is trained jointly with adversarial, feature matching and perceptual losses. Experimental results on the Cityscapes dataset demonstrated that the proposed PAN semantic segmentation model outperformed UNet, FCN and other baseline models in mIoU. The developed GAN reconstruction method produced visually pleasing images at low bit rates. Robustness tests under various channel conditions further verified that the proposed scheme maintained stable reconstruction performance.

\end{document}